\documentclass[runningheads]{llncs}
\usepackage[T1]{fontenc}
\usepackage{graphicx}
\usepackage{amssymb}
\usepackage{amsmath}
\usepackage{algorithm}
\usepackage{algorithmic}
\usepackage{xcolor} 
\usepackage{comment} 
\usepackage{hyperref}
\usepackage{tikz}
\usetikzlibrary{automata, positioning}

\newcommand\dd[1]{\textcolor{violet}{\textit{DD: #1}}}

\newcommand\oldtex[1]{\textcolor{red}{\textit{Old Text: #1}}}

\begin{document}
\title{Expressive Reward Synthesis\\ with the 
Runtime Monitoring Language} 
%
%

\author{Daniel Donnelly\inst{1} \and
Angelo Ferrando\inst{2} \and
Francesco Belardinelli\inst{1}}

\institute{
Imperial College London \and
University of Modena and Reggio Emilia}
%
%
%
 \maketitle              
%





\title{Expressive Runtime Verification\\ for Safe Reinforcement Learning}






\begin{abstract}
%
A key challenge in reinforcement learning (RL) is reward (mis)specification, whereby imprecisely defined reward functions can result in unintended, possibly harmful, behaviours. Indeed, reward functions in RL are typically treated as black-box mappings from state-action pairs to scalar values. While effective in many settings, this approach provides no information about why rewards are given, which can hinder learning and interpretability.  Reward Machines address this issue by representing reward functions as finite state automata, enabling the specification of structured, non-Markovian reward functions. However, their expressivity is typically bounded by regular languages, leaving them unable to capture more complex behaviours such as counting or parametrised conditions. In this work, we build on the Runtime Monitoring Language (RML) to develop a novel class of language-based Reward Machines. By leveraging the built-in memory of RML, our approach can specify reward functions for non-regular, non-Markovian tasks. We demonstrate the expressiveness of our approach through experiments, highlighting additional advantages in flexible event-handling and task specification over existing Reward Machine-based methods.
\end{abstract}


\section{Introduction}

Reinforcement Learning (RL)~\cite{sutton2018reinforcement} has achieved remarkable success by enabling agents to learn through interactions with their environment, using reward signals to shape their behaviour. Yet, the reward function that produces these signals is typically treated as a black box that the agent queries to receive rewards \cite{icarte2022reward}.


\emph{Reward Machines} (RMs)~\cite{icarte2022reward,icarte2018using} represent reward functions using finite state machines, enabling the agent to receive an explicit representation of the reward function. Each state in the machine corresponds to a possibly different reward function, with transitions between states triggered by events in the environment.


Furthermore, Reward Machines can encode histories of state-action sequences, allowing the specification of long-horizon objectives and multi-stage tasks. However, Reward Machines are typically limited to expressing non-Markovian properties that can be described by regular languages \cite{icarte2022reward}, thus making them unsuitable for tasks requiring more expressive capabilities, such as counting~\cite{bester2023counting} or parametrised conditions. 


This paper addresses these limitations by introducing \emph{RML Reward Machines}, which extend the expressivity of Reward Machines by leveraging the Runtime Monitoring Language (RML)~\cite{ancona2021rml}. Building on prior work on RMLGym~\cite{unniyankal2023rmlgym}, which first applied RML to reinforcement learning, our framework enables RML monitors to function as reward machines by providing the monitor state to the agent and introducing intermediate rewards. RML provides mechanisms for parametric event handling, allowing the specification and storage of complex properties in memory. 
These features enable tasks with memory requirements, such as counting or conditional behaviour based on past observations, to be encoded directly in the reward function. As a result, a broader range of tasks can be accurately specified and learned, supporting new use cases in domains such as robot navigation. Additionally, by allowing more precise task definition, our approach helps mitigate reward misspecification caused by underspecified objectives.

\paragraph{Our Contribution.}
We introduce RML Reward Machines, a novel logic-based approach to Reward Machines that leverages the expressive power of RML.
Our method enables agents to learn non-regular, non-Markovian tasks with memory-based objectives that traditional Reward Machines cannot capture. To achieve this, we build on the RMLGym framework ~\cite{unniyankal2023rmlgym}, by providing agents with a representation of the monitor state, allowing them to distinguish between different phases of the task and receive intermediate rewards. Empirical results demonstrate significant advantages in task specification and event handling compared to existing RM-based approaches and show that exposing the monitor state allows agents to learn more effectively on history-dependent tasks than agents trained with RMLGym. Code used to run the experiments is available at \url{https://github.com/danieldonnelly7/rml_reward_machines}.

\section{Background}

In this section, we provide the necessary background on reinforcement learning \cite{sutton2018reinforcement} and the Runtime Monitoring Language (RML) \cite{ancona2021rml} required to understand our approach.

\subsection{Reinforcement Learning}

Reinforcement learning allows agents to learn by interacting with an environment to develop optimal policies that maximise the expected sum of discounted rewards received over time. Throughout this work, a \emph{probability distribution} over a set \(X\) is defined as a function \(P: X \to [0,1]\) satisfying \(P(x) \geq 0\) for all \(x \in X\), and \(\sum_{x \in X} P(x) = 1\). 
We denote the set of all such probability distributions as \(\Delta X\).
Reinforcement learning problems are generally modelled as Markov Decision Processes.
\begin{definition}[MDP]
A \emph{Markov Decision Process} is 
a tuple $M = (S,A,T,R,\gamma)$, where (i)
$S$ is the finite set of states; (ii) $A$ is the set of actions; (iii) $T: S \times A \to \Delta S$ is the transition function; 
(iv) $R: S \times A \times S \to \mathbb{R}$ is the reward function; and (v) $\gamma \in (0,1]$ is the discount factor.
\end{definition}

A \emph{policy} $\pi: S \to \Delta A$ is a mapping from states to action distributions. For each state $s \in S$, actions are chosen according to a probability distribution over $A$, denoted as $\pi(a | s)$. As the agent interacts with the environment, they observe a trajectory of states, actions, and rewards, which is denoted as \newline{$\tau = (s_0, a_0, r_1, s_1, \dots, s_{n-1}, a_{n-1}, r_n, s_n)$}. The goal of the agent is to learn an \emph{optimal} policy $\pi^*$ that maximises the expected return.  
The return for a trajectory is defined as $G = \sum_{k=0}^{\infty} \gamma^k R_{k+1}$, and the optimal policy is defined as $\pi^* = \arg\max_{\pi} \mathbb{E}_\pi[G]$.

\vspace{-1em}
\subsection{Runtime Monitoring Language}

Runtime Verification (RV) is a lightweight approach for monitoring systems online by checking properties against the system's behaviour at runtime \cite{falcone2013tutorial}. 
In runtime verification, properties are verified over traces of events.
%
    A {\em finite trace} $\sigma \in EV^*$ is 
    a finite sequence $Ev_1Ev_2Ev_3 \ldots$ of events, where each $Ev_i$ comes from a possibly infinite set $EV$  of events that can be generated by the system (\textit{i.e.}, the system's alphabet).    

The \textit{Runtime Monitoring Language}\footnote{\url{https://rmlatdibris.github.io/}} (RML)~\cite{ancona2021rml} is a domain-specific language 
for specifying properties in RV, especially those requiring high expressiveness (e.g., non-context-free properties). We adopt RML in this work for its parametric capabilities.

%
%
%
%
The two components of an RML specification are {\em event types} and {\em terms}. Intuitively, the event types match events from the system and are used to construct RML terms. We introduce each of these components in the following definitions.

\begin{definition}[Event Type]
    An atomic event type $ET$ is a set of key-value pairs $\{k_1: v_1, \ldots, k_n: v_n\}$, where each key $k_i$ identifies specific information and $v_i$ is the matching condition. 
    The event type grammar follows:
\begin{align*}
ETs &::=  {ET}_1; \dots; {ET}_n \\
{ET} &::= \lambda(x_1, \dots, x_n) \text{ match } \text{op}\\ 
&\quad\ |\ \lambda(x_1, \dots, x_n) \neg \text{match } \text{op}\\
&\quad\ |\ \lambda(x_1, \dots, x_n) \text{ match } \text{etp}_1 \mid \dots \mid \text{etp}_n \\
&\quad\ |\ \lambda(x_1, \dots, x_n) \neg \text{match } \text{etp}_1 \mid \dots \mid \text{etp}_n \\
vp &::= x \mid l \mid op \mid ap \mid w\\
op &::= \{k_1:vp_1, \dots, k_n:vp_n\}\\
ap &::= [vp_1, \dots, vp_n] \mid [vp_1, \dots, vp_n, el]\\ 
etp &::= \lambda(vp_1, \dots, vp_n)
\end{align*}  
where \( \lambda \) denotes the event type name and the match statement specifies patterns for key-value pairs that an event must satisfy to match the event type. These patterns may include variables (\( x \)), primitive literals (\( l \)) such as numbers, strings, and booleans, as well as object patterns (\(op\)), array patterns (\(ap\)), and wildcards (\( w \)).
%
The ellipsis symbol (\( el \)) in array patterns enables matching arrays of variable length.
\end{definition}

\begin{definition}[Matching Event]\label{def:event}
    An \textit{event} $Ev$, also a set of key-value pairs, matches $ET$ if $ET \subseteq Ev$, i.e., for every $(k_i: v_i) \in ET$, there exists $(k_j: v_j) \in Ev$ such that $k_i = k_j$ and $v_i = v_j$.  
\end{definition}

Essentially, an event type specifies the criteria that an event must meet within the specification. 
Separating event types from the main specification enables complex matching logic to be defined independently of the high-level task description. This separation makes the specification simpler and more readable, as only the event type names appear in the RML term. Event types can also be reused across the specification, promoting modularity. Conceptually, event types serve as the building blocks of RML terms, analogous to atomic propositions in logic-based languages.




\begin{example}[Variables in Event Types]
\label{event_type_example}

RML event types can include variables to enable flexible and context-sensitive specifications. For example, consider the event type:

\[
move(x, y) \text{ match } \{\text{action}: \text{``move''}, \text{direction}: x, \text{distance}: y\}
\]
%
where $x$ and $y$ are variables representing the direction and distance of an agent's movement. This event type matches any event with action: `move', and specified `direction' and `distance'.
For instance, an event:
%
%
\[
Ev \text{ match } \{\text{action}: \text{``move''}, \text{direction}: \text{``north''}, \text{distance}: 3\}
\]

matches $move$ by binding $x = \text{``north''}$ and $y = 3$.

\end{example}

An RML \textit{term} $t$ defines how event types combine to form valid sequences or patterns using various operators. The atomic term $ET$ represents singleton traces containing any event $Ev$ that matches the event type $ET$. Sequential composition ($t_1  t_2$) denotes traces where a sequence from $t_1$ is followed by one from $t_2$. Unordered composition or {\em shuffle} ($t_1 \mid t_2$) allows traces from $t_1$ and $t_2$ to interleave, while preserving their internal order. Intersection ($t_1 \land t_2$) accepts traces satisfying both $t_1$ and $t_2$, whereas union ($t_1 \lor t_2$) accepts traces satisfying either. The Kleene star ($t^*$) denotes zero or more concatenations of $t$. The construct $\{ \text{let } x; \; t \}$ introduces a variable $x$ within $t$, enabling variables to appear in event types and unify with observed events. The full syntactic structure of RML terms is provided in Appendix A. 
We denote the set of all RML terms by $TE$.

Event types and RML terms are used together to create RML properties which represent the behaviour of the system being monitored.
%
\begin{definition}[RML Property]
    An {\em RML property} is 
    a pair $\langle t, \mathit{ETs} \rangle$, where $t$ is a term specifying the logical structure of event sequences, and $ \newline \mathit{ETs} = \{ET_1, \ldots, ET_n\}$ is a set of \textit{event types}. 
\end{definition}

\begin{example}[Variables and Parameters in Specifications]
\label{variables_and_param_example}
The event types and their bound variables from Example \ref{event_type_example} can be used as part of an RML specification. Variables allow RML specifications to enforce constraints on action sequences. For example, in the following specification, if an agent moves `north' by distance $y$, the next valid action must be `move' `south' by the same distance $y$. 
\begin{align}
Main = & \{\text{let } x, y; \ \text{move}(x, y) \nonumber \\ & \text{if } (x = \text{``north''}) \ \text{move}(\text{``south''}, y) \nonumber \\
 & \text{else } \text{move}(\text{``north''}, y) \}  \nonumber
\end{align}

Once variables are bound (e.g., $x = \text{``north''}$, $y = 3$), the specification can enforce subsequent events, such as returning ``south'' for the same distance $y$, ensuring valid behaviour according to the protocol.
This demonstrates the expressiveness of RML, allowing complex, parametrised rules and action sequences to be monitored and enforced, enabling nuanced control over the reward structure in reinforcement learning tasks.

\end{example}

When an RML term is compared to an event or trace of events, the system outputs a verdict that represents whether the term was satisfied by the event or trace. 
Although all traces observed at runtime are finite, RML defines verdicts by reasoning over their possible infinite continuations. This lets the system express whether the observed behaviour guarantees, precludes, or leaves open the possibility of satisfying the specification, based on all the ways the trace might evolve.
The set of verdicts $V$ used by RML contains four values which are defined as follows.
\begin{definition}[RML Verdicts] Let $T$ denote the possibly infinite set of traces generated by the system, and let $VT$ denote the set of all finite traces that match the RML specification. Given a current finite trace $\sigma$, its possibly infinite continuations $\sigma'$ (\textit{i.e.}, 
$\sigma$ is prefix of $\sigma'$),  the verdict $v \in V$ is defined as follows:
\begin{align*} 
v &= \text{True} \text{ iff }\sigma \in VT \text{ and for all }  \sigma' \in T,  \sigma' \in VT. \\ 
v &= \text{Currently True}  \text{ iff } \sigma \in VT \text{ and for some } \sigma' \in T, \sigma' \notin VT. \\ 
v &= \text{Currently False}  \text{ iff } \sigma \notin VT \text{ and for some } \sigma' \in T, \sigma' \in VT. \\ 
v &= \text{False}  \text{ iff } \sigma \notin VT \text{ and for all } \sigma' \in T, \sigma' \notin VT. \end{align*} \end{definition}

\section{RML Reward Machines}

To enable the use of memory-aware reward functions in RL, we introduce RML Reward Machines, a novel type of reward machine that leverages the expressive power of RML. This section begins by adapting the RML formalism for compatibility with RL notation (Sec.~\ref{sec:ext_formalism}). We then present the RML Reward Machine framework (Sec.~\ref{sec:rmlrm}), detailing its integration with MDPs and the mechanisms that allow agents to leverage memory encoded in monitor states to improve decision-making (Sec.~\ref{sec:expressivity}).


\subsection{Extended RML Formalism} \label{sec:ext_formalism}



A system connected to an RML monitor includes instrumentation that processes events into a trace compatible with the RML monitor. The events in the trace are processed sequentially and are first compared against the event types. This matching process is presented in Definition~\ref{def:event}, and is used to generate the set of event types that match a given event. This process can be described by a function $L:EV\rightarrow 2^{ETs}$ which maps any event to a set of matched event types $M$, \textit{i.e.}, $L(Ev_i)=M$. After this matching process, $M$ can be compared with the RML term $t$.

The matched event types are compared with the corresponding event types at the current state of the RML term. 
The current state could be a single event type or a more complex logical condition, such as an intersection, which requires the event to match each of the constituent event types.
After the comparison is finished, the RML term advances to the next element in the term, which the next set of matched event types in the trace is compared to. The progression to the next element is determined by the operational semantics of RML. A full description of the operational semantics is outside the scope of this paper but can be found in \cite{ancona2021rml}. For our purposes, we define the operational semantics of RML in a functional manner.
%
\begin{definition}[Functional Definition of Operational Semantics]
Let $t, t' \in TE$ be RML terms, $ETs_{all}$ be the set of all possible event types, and let $K \subseteq ETs_{all}$ represent a subset of event types matched by an event. The operational semantics of RML is described by a function, $\delta: TE \times 2^{ETs_{all}} \rightarrow TE$, where $\delta(t,K)=t'$, indicates that a term $t$ transforms into $t'$ upon observing the set of event types $K$.
\label{functional_def_op_semantics}
\end{definition}


After an event is processed against a given term, the term changes to a new variant. If it is possible to transform to a term $t'$ from an initial term $t$, we say that $t$ is \emph{reachable}, which can be defined more formally as follows:
\begin{definition}[Reachability] \label{reachability}
%
An RML term $t'$ is said to be \emph{reachable} from an initial RML term $t$ if there exist a sequence $Ev_1, \ldots, Ev_i, \ldots, Ev_n$ of events and intermediate terms $t_i$, $1 \leq i < n$, such that $t_1 = t$; $t_n = t'$; and for $1 \leq i < n$, $t_i \xrightarrow{Ev_i} t_{i+1}$, where \( \xrightarrow{Ev^i} \) denotes the operational semantics of RML.


We denote this reachability relation as $\xrightarrow{Ev^*}$, and the set of terms reachable from an RML term $t$ as $W = \{ t' \mid t \xrightarrow{Ev^*} t' \}$.
\end{definition}

Each time an event is processed the RML monitor comes to a verdict. This process can be represented as a function $\delta_v:W \times 2^{ETs} \rightarrow V$ with $\delta_v(t',M)=v$. 
The term $t'$ encodes the history of events in the trace $\sigma$. Therefore, when we evaluate $\delta_v(t', M)$, the monitor's verdict reflects the influence of the entire event history up to the current point.

\begin{example}[Conditional RML Specification]

A specification for an ordered sequence of events with an if-else conditional operator can be given as follows:
\begin{quote}
    \centering
    a matches \{event : `a'\}; \\
    b(n) matches \{event : `b', val : n\}; \\
    c matches \{event : `c'\}; \\
    d matches \{event : `d'\}; \\
    \vspace{1em}
    Main = a \{let n; b(n) if (n > 2) c else d\};
\end{quote} 

Let a trace of events be defined as $\sigma = Ev_1 Ev_2 Ev_3$, where: $Ev_1 = $ \{event : `a'\}, $Ev_2 = $ \{event : `b', val : 3\}, $Ev_3 = $ \{event : `c'\}. Let an initial RML term be denoted by:
%
\begin{quote}
    \centering
    $t_0=a \{let \text{ n}; \space b(n) \text{ if } (n > 2) \space c \text{ else } d\}$
\end{quote}

The events are processed by the matching function to get the matched event types with $L(Ev_1)=$ \{a\}, $L(Ev_2)=$ \{b(3)\}, and $L(Ev_3)=$ \{c\}.

The matched event types for $Ev_1$ contain an `a' event which matches the first part of the RML term, leading to the update:
\begin{quote}
    \centering
    $\delta(t_0,\{a\}) = t_1 = \{let \text{ n}; \space b(n) \text{ if } (n > 2) \space c \text{ else } d\}$
\end{quote}

At the same time, the verdict is determined as $\delta_v(t_0,\{a\}) = $ \emph{Currently False}, as the whole specification is not satisfied, but further sequences of events could lead to the specification being satisfied. Following this $Ev_2$ is observed, which results in the set of matched event types \{b(3)\}. Here, the observed variable `val' is bound as a parameter. This observation results in the new RML term $\delta(t_1,\{b(3)\}) = t_2 = c$. As $n$ is bound as 3, this makes the condition true, causing the specification to transform in line with the condition logic. If the value of $n$ had instead been 2, the condition would be false, and would have resulted in $t_2 = d$. The verdict after this comparison is again $\delta_v(t_1,\{b\}) = $ \emph{Currently False}, for the same reason as before. The final event in the trace $Ev_3$ with the corresponding matched event type set \{c\} leads the term to update to $\delta(t_2,\{c\})= t_3$ where $t_3$ denotes the \emph{empty} specification, which no sequence of events can satisfy. This final event leads to the verdict $\delta_v(t_2,\{c\})= $ \emph{Currently True}, as the specification is currently satisfied.
However, further events would make the full trace of events invalid.

\end{example}

\subsection{Definition of RML Reward Machines}
\label{sec:rmlrm}

\begin{figure}[ht]
\centering
\includegraphics[width=.7\linewidth]{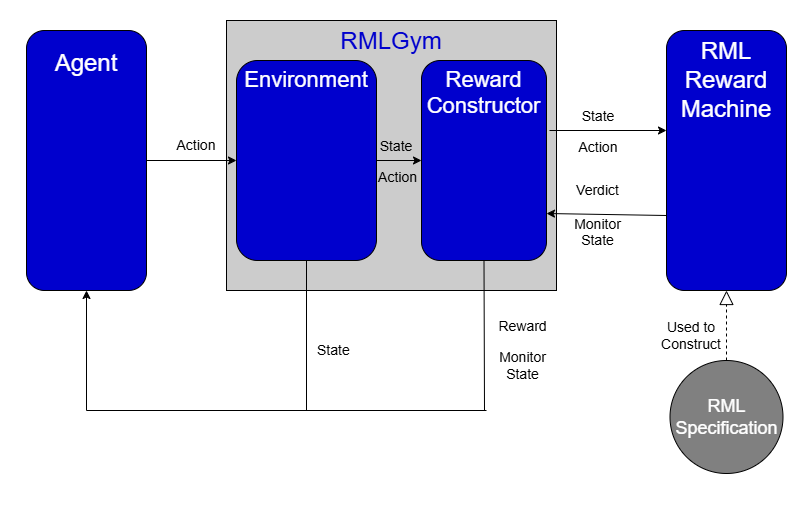}
\caption{RML Reward Machine Framework.}
\label{fig:rmlgym_diagram_new}
\end{figure}

RML Reward Machines are a highly expressive approach to language-based 
monitoring. 
They are connected to an MDP through a two-way communication channel, as shown in Figure \ref{fig:rmlgym_diagram_new}. This builds on the design used by the RMLGym framework \cite{unniyankal2023rmlgym}, where RML monitors receive a trace and send a verdict back to the system. 
One key shortcoming of the RMLGym framework is that rewards can appear non-deterministic from the agent's perspective since, for a given state-action pair, the agent may receive different rewards depending on the internal state of the RML monitor, which is invisible to the agent in RMLGym.
RML Reward Machines address this issue by sending the monitor state back to the system, providing the additional context required to make rewards deterministic from the agent’s viewpoint.
The information communicated back to the system passes through the Reward Constructor, which acts as an intermediary between the machine and the environment and is where the reward is computed. Note that the RML Reward Machine operates as a runtime controller that augments the agent's state with additional task-specific memory. While this introduces two‑way communication with the environment and adds latency in our prototype, much of the cost could be eliminated through optimisation. The full connected learning system is referred to as an RML-extended MDP and defined below. 



\begin{definition}[RML-extended MDP]
Let $M = (S, A, T, \delta_v, R, \gamma)$ be an MDP and $(t, ETs)$ an RML property. 
An {\em RML-extended MDP} is defined as a tuple \( \Gamma = (ETs, t, S \times W, A, T', \delta_v, R, \gamma) \), where (i) 
%
$A$ and  $\gamma$ are defined as in $M$;
%
(ii) $ETs$ and $t$ are defined as in the RML property. Moreover


\begin{itemize}

\item[(iii)] The state space \( S \times W \) is defined as the  Cartesian product of the MDP state space $S$ 
and set \( W \) of all reachable variants of \( t \) (as defined in Def.~\ref{reachability}). 



\item[(iv)] The transition function \( T' : (S \times W) \times A \rightarrow S \times W  \) maps a state-action pair to a new state, and is given by \( T'((s, t'), a) = (T(s'|s, a), \delta(t', L((s, a)))) \), where \( L \) and \( \delta \) are as defined in Section \ref{sec:ext_formalism} and Definition \ref{functional_def_op_semantics}. 

\item[(v)] The verdict function \( \delta_v: (S \times W) \times A \rightarrow V \) assigns a verdict based on the current state and action, where \( V \)  denotes the set of verdicts. Specifically, \( \delta_v(t', L((s, a))) \) determines the verdict from the RML term and matched event types. 

\item[(vi)] The reward function \( R: V \times W \times (S \times W) \rightarrow \mathbb{R} \) maps a verdict, the RML term, and the current state to a real-valued reward. 

\end{itemize}
\end{definition}


Algorithm \ref{alg:rml_reward_machine} describes how an input event is processed by the RML Reward Machine framework.
The event is compared against the event types, and the resulting matched event types are checked against the current RML specification. The specification then updates, represented by $\delta(t,L(Ev)) = t'$. The monitor also outputs the verdict via $\delta_v(t,L(Ev)) = v$, where $v \in V$. The verdict $v$ and new monitor state $t'$ are communicated back to the system, and are used as inputs to the reward function. Additionally, $t'$ is given to the agent as part of the state, along with the environment state. 

\begin{algorithm}
\caption{RML Reward Machine Update Procedure}
\label{alg:rml_reward_machine}
\begin{algorithmic}[1]
\STATE \textbf{Input:} Event $Ev = (s,a)$ containing environment state $s$ and action $a$; Current monitor state $t'$
\STATE \textbf{Output:} Updated state $(s,t'')$; Verdict $v$
\STATE Match events in $Ev$ with event types $ETs$, to compute $L(s,a)$
\STATE Update the monitor state: $t'' \gets \delta(t', L(s,a))$
\STATE Compute the verdict: $v \gets \delta_v(t', L(s,a))$
\STATE Communicate monitor state $t''$ and verdict $v$ back to the system
\STATE Compute reward $r \in \mathbb{R}$: $r \gets R(v,t'',(s,t''))$
\STATE State $(s,t'')$ and reward $r$ communicated to the agent.
\RETURN $(s,t''), r$
\end{algorithmic}
\end{algorithm}


\paragraph{Discussion.} The RML formula is an ordering of events, connected by various operators, representing the task the agent is expected to learn. This formula can be abstracted into a state machine representation, resembling a Reward Machine \cite{icarte2018using,icarte2022reward}, but with the distinction of potentially having an infinite number of states, due to RML's ability to store variables. In this representation, the constituent elements in the RML specification term correspond to machine states. Elements that incorporate memory can be represented by unique states or memory variables such as counters. Transitions to new elements in the formula can be represented by edges between machine states. Observations that cause a False verdict cause a transition to a terminal failure state, while observations that cause a True or Currently True verdict transition to a terminal success state. 
Other observations keep the system in the same state or transition to non-terminal states. Rewards are also based on the current term of the RML formula. 
The total number of states in an RML-extended MDP is given by $|S|\times |W|$, where $|W|$ may be infinite in tasks that recursively expand or require unbounded memory. 

The monitor state is communicated from the monitor to the system. The system leverages this state in two main ways: 
\begin{enumerate}
    \item The state $s' \in S \times W$ 
    visible to the agent, is obtained by providing the monitor state $t' \in W$ as part of a cross-product with the environment state $s \in S$, resulting in $s'=(s,t')$. Without this information rewards can be non-deterministic from the perspective of the agent, as the same environment state-action pair can be associated with different rewards depending on the monitor state. This approach ensures that rewards are deterministic from the agent's perspective.
    
    \item The integration of the monitor state $t'$ into the system enables its use for reward specification. This is in addition to a reward based on the verdict of the monitor, with each verdict having an associated reward. In multi-stage tasks, the verdict-based reward process typically does not provide a reward for advancing to the next stage of a task. This limitation can be overcome using the RML term, by granting a reward when the RML term transitions to a new variant of the term, i.e., $\delta(t',L(Ev)) = t''$ with $t' \neq t''$. This process can be viewed as an automated form of reward shaping \cite{ng1999policy}, where monitor state transitions signal progress toward the final objective. An optional additional reward process can be used to encourage exploration by providing a new reward each time a new state (environment and monitor state combination) is observed.
\end{enumerate}

Because an RML monitor can store unbounded counters or arbitrary data values, the set \(W\) of monitor states can be countably infinite for certain problems. Thus, in the general case, standard tabular reinforcement learning algorithms are not guaranteed to converge \cite{sutton2018reinforcement}. A potential solution is to create a finite representation of the state space, for instance, by representing monitor states using a one-hot vector, assigning each memory parameter an index, and marking active memory values with a 1 at the corresponding index. However, this may lose critical information required for a task. Alternatively, when memory parameters are numeric, their value can be encoded directly on the vector, providing a more suitable representation for function approximation-based methods. While this can improve learning performance, it does not eliminate the underlying issue of an infinite state space, and convergence guarantees remain absent. In our experiments, we use settings that require a finite number of monitor states, making standard tabular RL algorithms suitable.


\subsection{Expressivity and Flexibility of RML Reward Machines} \label{sec:expressivity}

\paragraph{Expressivity.}


Reward Machines, as a type of deterministic finite automaton, can express non-Markovian reward functions over state-action histories corresponding to regular languages. Unlike standard Reward Machines, which are limited to this class, RML Reward Machines extend expressivity by supporting memory, variables, and parametric event handling.
This extended expressivity enables RML Reward Machines to specify non-Markovian reward functions that lie beyond the regular language class -- for example, tasks involving counting. Moreover, since RML supports all operators found in regular expressions~\cite{ancona2021rml} (e.g., concatenation, union, and Kleene star), any regular language -- and by extension, any reward function definable by a standard Reward Machine -- can be encoded as an RML monitor. As such, RML Reward Machines strictly generalise standard Reward Machines in terms of expressiveness. Beyond regular languages, RML is also more expressive than LTL under three-valued semantics over finite traces, as it builds upon and extends trace expressions -- a formalism that has been formally shown to surpass LTL in expressive power~\cite{ancona2016comparing}. This expressiveness derives from RML’s ability to store and reason over variable bindings, support conditional logic, and match complex event structures, enabling it to specify behaviours and reward conditions not representable in LTL or traditional reward specification frameworks. However, a precise automata-theoretic characterisation of RML monitors -- particularly regarding closure properties, expressiveness classes, and decidability -- remains an open question, beyond the scope of the present contribution.




An example that showcases the counting property is a reward function that grants rewards if it observes any string where an event $A$ is observed $N$ times followed by $N$ occurrences of an event $B$, which can be described by the set $\{A^{N}B^{N} : N \in \mathbb{N}\}$. For any individual value of $N$, a Reward Machine can be constructed to represent this task. However, a Reward Machine cannot be designed that can represent this task for all $N \in \mathbb{N}$. In contrast, an RML formula can be constructed to represent this task, as shown in Figure \ref{fig:coutning_RML_formula}. The formula utilises the generic layer of RML, allowing the value of $N$ to be instantiated and stored within the definition. When $A$ is observed (represented by $a$), the stored parameter $n$ in the $A$\textless $n$\textgreater definition increases by 1. This happens until $B$ is observed (represented by $b$), at which point $n$ is decremented by 1, and the new value $n-1$ is stored in the $B$\textless $n$\textgreater definition. Following this, $b$ is repeatedly observed until the count reaches 0, at which point the formula concludes and the task is completed.
\begin{figure}[ht]
\centering
\begin{quote}
    \centering
    Main = $A$\textless1\textgreater ;  \\
    $A$\textless $n$\textgreater = $a$ ($A$\textless $n+1$\textgreater $\vee$ $B$\textless $n-1$\textgreater); \\
    $B$\textless $n$\textgreater = if ($n$\textgreater 0)   $b$ $B$\textless $n-1$\textgreater \space\space else $b$;
\end{quote} 
\caption{RML Formula Counting Example}
\label{fig:coutning_RML_formula}
\end{figure}
\vspace{-1em}

\paragraph{Flexibility.}
A second notable strength of RML Reward Machines is their flexibility in event handling. All transitions in a Reward Machine normally need to be pre-specified, including what events the transition occurs in response to. RML Reward Machines on the other hand  only require the event to be formatted with the correct structure to match event types which contain variables. These matched values are bound and can be used later in the specification, informing the sequence of events the agent is required to perform. This is particularly useful for numerical tasks, where a number is given that corresponds to an event. Earlier in Example~\ref{event_type_example} and 
\ref{variables_and_param_example} a specification for such a numerical task was given. The definition of the event type matches any distance $y$ provided the action is `move' and a direction is given. When this event is observed, the specification binds the value of $y$ for use later on.

Pre-specifying Reward Machines for numerical tasks with large value ranges can require many defined states and transitions. Each value must be addressed by a state or transition, and observations outside the defined range cannot be processed. To match the flexibility of the specification in Example~\ref{variables_and_param_example}, each possible combination of $y$ and direction must be represented by a state. As a result, specifying the machine becomes increasingly complex as $y$ grows.



\section{Experimental Evaluation}

In the experiments, we utilise tabular Q-learning \cite{watkins1992q}, a model-free method that directly estimates the Q function. We employ the $\epsilon$-greedy policy throughout, which selects a random action with probability $\epsilon$ and the action with the highest expected return with probability $1- \epsilon$. 

\subsection{Environment}

\begin{figure}[ht]
\centering
\includegraphics[width=.5\linewidth]{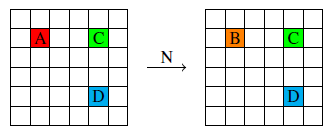}
\caption{The LetterEnv Environment (from \cite{bester2023counting}).}
\label{fig:letter_env}
\end{figure}

The experiments in this section use variations of the LetterEnv environment, shown in Figure~\ref{fig:letter_env} 
The environment is a grid with letters positioned on its squares. Tasks in this environment involve observing a specific sequence of the letters on the board. The task illustrated in Figure~\ref{fig:letter_env} involves following the sequence $\{A^NBCD^N : N \in \mathbb{N}\}$. The letters on the grid can be replaced by other letters after a specified number of observations. In the example, after $A$ is observed $N$ times it is replaced by $B$. 
If a letter is observed out of sequence the task is failed. Standard Reward Machines can only learn this task for specific values of $N$ \cite{bester2023counting}, with a general solution requiring a more expressive formalism.

\subsection{Numerical Experiment}

\begin{figure}[h]
\begin{quote}
\raggedright
a\_match(n) \textbf{matches} \{a : n\}; \\
b\_match \textbf{matches} \{b : t\} with t = 1.0; \\
c\_match \textbf{matches} \{c : t\} with t = 1.0; \\
d\_match \textbf{matches} \{d : t\} with t = 1.0; \\
not\_abcd \textbf{not matches} \quad a\_match \texttt{|} b\_match \texttt{|}  c\_match \texttt{|}  d\_match; \\
\vspace{1em}

Main = not\_abcd* \{let n; a\_match(n) not\_abcd* B\textless n\textgreater\}; \\
B\textless n\textgreater = b\_match C\textless n\textgreater; \\
C\textless n\textgreater = not\_abcd* c\_match D\textless n\textgreater; \\
D\textless n\textgreater = if (n \textgreater 0) not\_abcd* d\_match D\textless n-1\textgreater \quad else all;
\end{quote} 
\caption{Numerical Experiment RML Formula}
\label{fig:numerical_experiment_rml_formula}
\end{figure}

In this section, the standard LetterEnv environment, shown in Figure \ref{fig:letter_env} was modified so that the letter $A$ outputs a number instead of the letter. In this experiment, the letter $A$ is observed only once, with the output number corresponding to $N$, the number of times A would normally be observed. After observing $A$, the agent is tasked with observing $B$, then $C$, and finally observing $D$, which must be observed $N$ times. The full string expected to be observed is 
\{$A(N)BCD^{N}$ : $N \in \{1,2,3,4,5,6,7,8,9,10\}\}$.  For the purposes of the experiment, we limited $N$ to a finite range up to 10. The RML specification for this task is shown in Figure \ref{fig:numerical_experiment_rml_formula}.

\subsubsection{Flexible Event Handling}
\label{Flexible_event_handling_section}

In this experiment, RML Reward Machines are compared against two versions of Counting Reward Automata (CRA): using Q-Learning (QL) and using Counterfactual Q-Learning (CQL) \cite{bester2023counting}. Counting Reward Automata 
are chosen for comparison as they are the only other RM-based approach that leverages memory, in the form of counters. Full experimental details can be found in the Appendix B.1. 



\paragraph{Results.}

\begin{figure}[ht]
\centering
\includegraphics[width=0.8\linewidth]{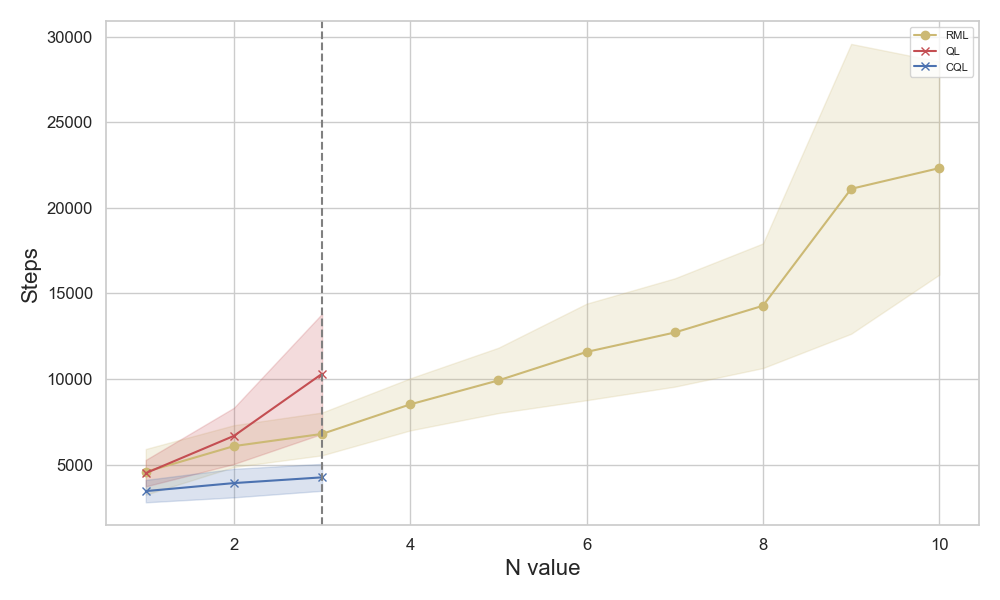}
\caption{Numerical Inputs Flexibility Experiment Results. Mean result and 1 standard deviation interval shown in shaded region. Yellow = RML Reward Machines, Red = QL, Blue = CQL}
\label{fig:num_inputs_experiment}
\end{figure}

The results of the experiment are shown in Fig.~\ref{fig:num_inputs_experiment}. For the values of $N$ that all approaches can handle, RML Reward Machines learn faster than QL and slower than CQL, demonstrating how counterfactual learning can accelerate the learning process. 

When $N$ is greater than 3, the CRA-based approaches fail at the task. This is because RM-based approaches normally require each observable event to be explicitly defined. The machines in this case were only designed to handle values up to 3, and could not perform the task for higher values of $N$, as those values were not defined as observable events. While $N=3$ was chosen arbitrarily for this experiment, the same limitation applies for any predefined threshold. Additionally, as this threshold grows larger, the task of specifying the counting reward automata becomes more complex, as each additional input needs to be defined. 

RML Reward Machines, on the other hand, successfully learn the task for all tested values of $N$. This is possible because RML supports the use of variables in event type definitions. These variables can take on any value, allowing the machine to process all possible values of $N$, rather than being restricted to a predefined range.


\subsubsection{Effect of Monitor State Visibility}

To demonstrate the effect of making the monitor state visible to the agent, RML Reward Machines are compared against RMLGym \cite{unniyankal2023rmlgym} using the same numerical LetterEnv setup as in the previous experiment. Without the monitor state, rewards can become non-deterministic from the perspective of the agent. For example, an agent positioned on a square adjacent to the letter $C$ may receive different rewards when moving to $C$, depending on the unobserved history of events. If $A(N)$ and $B$ have already been observed, moving to $C$ is the correct next step and is on the path to receiving a positive reward. Conversely, if that sequence has not been observed, the same action leads to failure and a negative reward signal.

For this experiment, the value of $N$ is fixed at $1$. The RML Reward Machine setup described in Section~\ref{Flexible_event_handling_section} is compared against RMLGym. In addition, we include an ablated version of the RML Reward Machine that receives no intermediate rewards when the RML term transitions between variants, receiving rewards only upon verdicts. All approaches use the same task specification as in the previous experiment (Figure~\ref{fig:num_inputs_experiment}). The methods are trained for 1000 episodes, and performance is evaluated using the number of successful task completions over a rolling 50-episode window. Full experimental details are provided in Appendix~B.2. 

\begin{figure}[ht]
\centering
\includegraphics[width=0.9\linewidth]{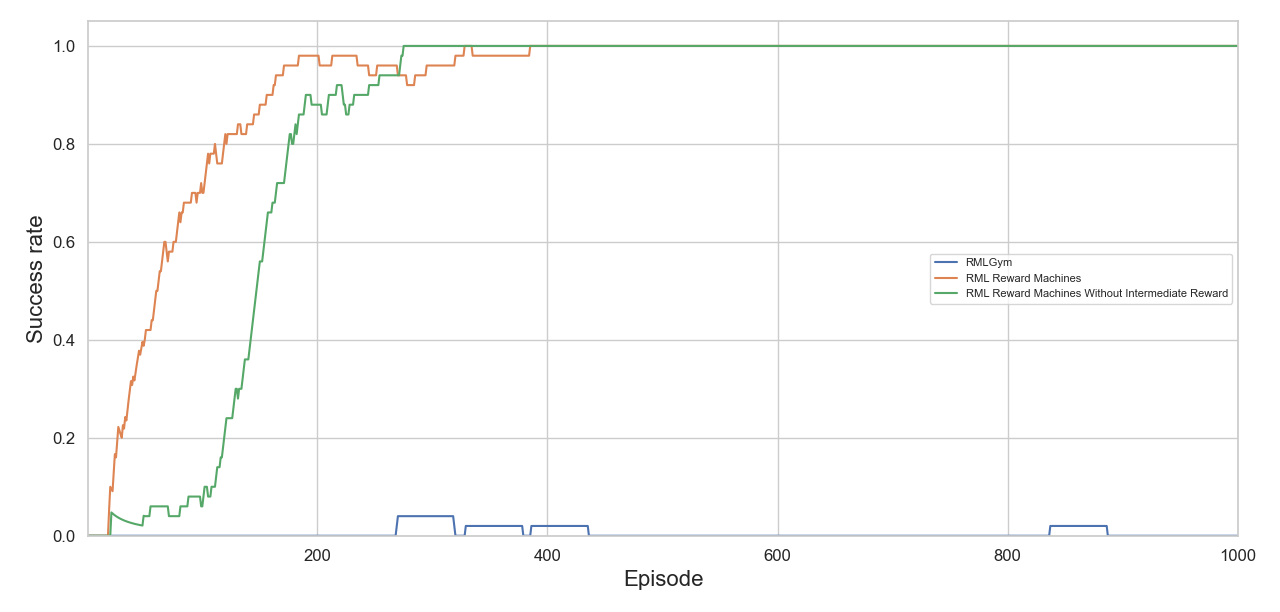}
\caption{Numerical Inputs Monitor State Experiment Results. Success rate over the last 50 episodes. Orange = RML Reward Machines, Green = RML Reward Machines (ablated), Blue = RMLGym}
\label{fig:RMLGym_comparison_results}
\end{figure}

\paragraph{Results.}

The results of the comparison with RMLGym are shown in Fig.~\ref{fig:RMLGym_comparison_results}. RML Reward Machines successfully learn the task both with and without intermediate rewards. The inclusion of intermediate rewards generally accelerates learning, although the ablated version converges to an always-successful policy slightly earlier. Since both approaches stabilise at approximately the same time, we hypothesise that this difference is due to errors from random actions, which diminish as $\epsilon$ decreases. RMLGym, on the other hand, fails to learn the task reliably, indicating that the absence of monitor-state information impedes learning.

\subsection{Complexity Analysis of Conditional Tasks}


RML Reward Machines allow for seamless specification of a range of non-regular properties. In this section, we demonstrate this feature using parametric conditional tasks, which require different behaviour conditional on the value of a parameter. To compare ease of specification, we measure specification complexity by counting the number of explicit branches needed to handle each of the possible outcomes. By leveraging the ability of RML to store parametric values and use the if-else operator, RML Reward Machines can represent such tasks with a constant number of branches in the size of the specification.
In contrast, existing reward machine-based approaches require a linear number of 
branches.

\paragraph{Problem Setup.}
The LetterEnv environment is set up in its default format, where A is observed $N$ times followed by being replaced by B, with C and D also present on the grid. In this case the task depends on the value of $N$ and a second value $M$, which is a set constant value. A is observed $N$ times, followed by observing B. After this, the next observation depends on how many times A was observed in total. If A was observed less than $M$ times, the next observation should be C; otherwise, the next observation should be D. The two potential task strings are as follows:
\[
\left\{
\begin{array}{ll}
A^N BC & \text{if } N < M, \\
A^N BD & \text{if } N \geq M
\end{array}
\right.
\]

The RML specification for the task is shown in Figure~\ref{fig:conditional_experiment_rml_formula}, with $M$ set to 3.

\begin{figure}[h]
\begin{quote}
\raggedright
a\_match \textbf{matches} \{a : t\} with t = 1.0; \\
b\_match \textbf{matches} \{b : t\} with t = 1.0; \\
c\_match \textbf{matches} \{c : t\} with t = 1.0; \\
d\_match \textbf{matches} \{d : t\} with t = 1.0; \\
not\_abcd \textbf{not matches} \quad a\_match \texttt{|} b\_match \texttt{|}  c\_match \texttt{|}  d\_match; \\
\vspace{1em}

Main = not\_abcd* A\textless 0\textgreater; \\
A\textless n\textgreater = a\_match not\_abcd* (A\textless n+1\textgreater $\vee$ B\textless n+1\textgreater); \\
B\textless n\textgreater = b\_match C\textless n\textgreater; \\
C\textless n\textgreater = if (n \textgreater 2.5) not\_abcd* c\_match else not\_abcd* d\_match;
\end{quote} 
\caption{Conditional Experiment RML Formula}
\label{fig:conditional_experiment_rml_formula}
\end{figure}

\begin{figure}[ht]
\centering
\includegraphics[width=.75\linewidth]{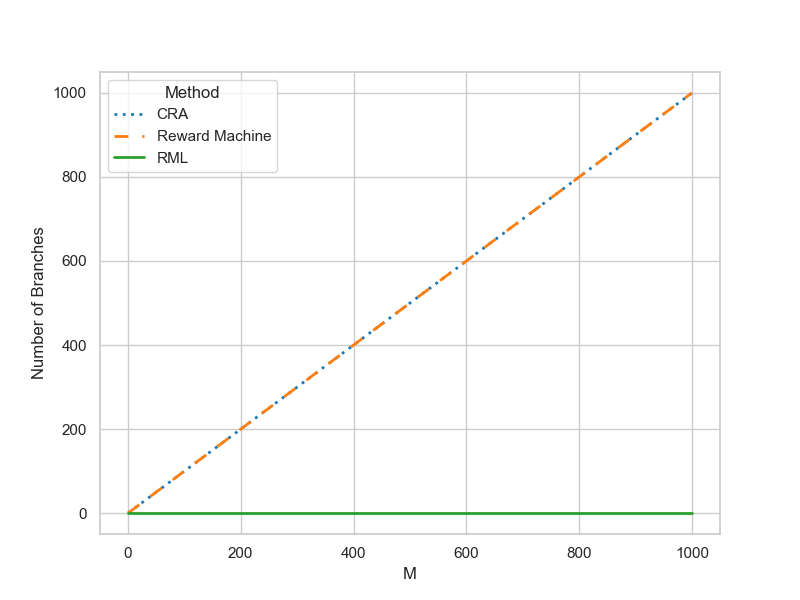}
\caption{Number of branches required for different values of environment condition value $M$. Reward Machines, CRA and RML reward Machines are represented by the orange dashed, blue dotted, and green lines respectively.
}
\label{fig:memory_parameters_graph}
\end{figure}

This task compares RML Reward Machines with standard Reward Machines and CRA, focusing on the complexity of task specification. We measure complexity by the number of \textit{branches} required to encode the task, where a branch is defined as a distinct automaton state or counter that explicitly represents a different value of $N$. Figure \ref{fig:memory_parameters_graph} displays the required number of branches to perform the task as $M$ increases.




\vspace{-1em}

\paragraph{Discussion}

For standard Reward Machines, the specification complexity grows linearly with $M$. Because Reward Machines cannot store $M$ directly, an explicit state is required to track each value of the count $N<M$. Once $N\ge M$, the machine can leverage a single state, as the behaviour remains the same for each possible $N$ value. In total, $M$ distinct branches are required, yielding an overall complexity of $\mathcal{O}(M)$.


Similarly to Reward Machines, the specification complexity grows linearly with $M$ for CRA. There are two potential approaches for representing this task using a CRA. Since CRA can represent any Reward Machine, the same state-based construction can be used, with complexity $\mathcal{O}(M)$. Alternatively, CRA can leverage their counters to store the value of $N$. However, CRA transitions can only be based on whether a counter is zero or non-zero. As such, for $ M > 1 $, the value of the counter cannot be used directly to specify transitions. Instead, $M$ counters, one for each possible value of $N$, have to be used to record the count by incrementing them by one in sequence. This approach mirrors the state-based approach used by standard Reward Machines, substituting states for counters. Similarly to the state-based approach, this approach requires $M$ explicit branches, giving overall complexity $\mathcal{O}(M)$.


RML Reward Machines avoid the need for an explicit branching structure, as stored values and conditions can be written directly in the specification. A single memory parameter can be used to store the value of $N$, which is incremented by one upon each observation of A. The if-else conditional operator can then be used to compare $N$ directly against $M$. Because the comparison outcome is used to select between the two possible continuations of the task, transitions for each value of $N$ do not need to be manually defined. As such, the task only requires a single branch up to the comparison, at which point two branches are introduced corresponding to the two potential continuations. Hence, the specification size remains constant as $M$ increases, with a resultant specification complexity of $\mathcal{O}(1)$.


\section{Related Work}

RML Reward Machines aim to address a limitation in the expressivity of standard Reward Machines \cite{icarte2018using,icarte2022reward} by leveraging memory. Reward Machines enable the specification of non-Markovian rewards by representing reward functions using finite-state automata. This framework allows for the precise definition of long-horizon, multi-stage tasks, mitigating the risk of reward misspecification that can occur when standard Markovian reward functions are used for such tasks. Extensions such as Numeric Reward Machines~\cite{levina2024numeric} add quantitative reasoning, while Pushdown Reward Machines \cite{varricchione2025pushdown} add stack-based memory to enhance expressiveness. Counting Reward Automata (CRA)~\cite{bester2023counting} provides memory via counters, achieving Turing-completeness when two or more counters are used. While CRA also utilise memory, they are limited to Boolean observations, whereas RML Reward Machines support parametric specifications, enabling richer data handling (e.g., strings and numeric values). Additionally, RML avoids manual automata construction, simplifying task specification. Temporally extended tasks have also been addressed using hierarchical RL frameworks such as HAMs \cite{parr1997reinforcement}, the options framework \cite{sutton1999between}, and MAXQ \cite{dietterich2000hierarchical}. 

A variety of language-based methods for reward specification have been developed, many of which incorporate variants of temporal logic \cite{aksaray2016q,sadigh2014learning,fu2014probably,li2017reinforcement}. 
Restraining bolts \cite{de2020restraining} leverage LTL over finite traces (LTL\textsubscript{f}) and its extension LDL\textsubscript{f}, to produce an external reward signal that encourages the agent to learn behaviours aligned with the given specifications.
SPECTRL \cite{jothimurugan2019composable} adopts a language-based approach that utilises quantitative information from the task. However, unlike RML Reward Machines, which use this information for task specification, SPECTRL applies it for reward shaping.

The RMLGym framework~\cite{unniyankal2023rmlgym} demonstrated RML's potential application to reinforcement learning by leveraging runtime monitors to provide rewards. This work builds on RMLGym by introducing intermediate rewards and exposing monitor states to the agent, accelerating learning and expanding the scope of expressible objectives.

\section{Conclusions and Future Work}

This paper introduced \textit{RML Reward Machines}, a novel framework that extends traditional Reward Machines by leveraging the expressiveness of RML. Key aspects of the framework include providing monitor states to the agent, introducing intermediate rewards, and enabling the specification of non-Markovian, non-regular reward functions that require memory. Empirical results demonstrated advantages in task specification and event handling over another memory-based Reward Machine approach, Counting Reward Automata, and showed that exposing the monitor state allows agents to learn more effectively on history-dependent tasks than agents trained using RMLGym.

Despite these advances, several avenues for future research remain. Leveraging counterfactual experiences during training for RML Reward Machines could enhance learning speed, as demonstrated by the improved performance of Counting Reward Automata in our experiments. Improved monitor state handling, for instance by leveraging vectors and sequence models, may speed up learning, while avoiding potential non-determinism in rewards. Evaluating the framework in safety-related (e.g., AI Safety Gridworlds~\cite{leike2017ai}) and high-dimensional environments requiring deep reinforcement learning would strengthen its practical applicability. Finally, a formal expressiveness analysis of RML would clarify its exact expressivity relative to other frameworks, such as Counting Reward Automata.



\paragraph{Acknowledgments.} The research described in this paper was partially supported by the EPSRC (grant number EP/X015823/1)  and by the Moro-Barry family.

\newpage

\bibliography{main}

\begin{thebibliography}{10}
\providecommand{\url}[1]{\texttt{#1}}
\providecommand{\urlprefix}{URL }
\providecommand{\doi}[1]{https://doi.org/#1}

\bibitem{aksaray2016q}
Aksaray, D., Jones, A., Kong, Z., Schwager, M., Belta, C.: Q-{L}earning for robust satisfaction of signal temporal logic specifications. In: 2016 IEEE 55th Conference on Decision and Control (CDC). pp. 6565--6570. IEEE (2016)

\bibitem{ancona2016comparing}
Ancona, D., Ferrando, A., Mascardi, V.: Comparing trace expressions and linear temporal logic for runtime verification. Theory and practice of formal methods: Essays dedicated to Frank de Boer on the occasion of his 60th birthday pp. 47--64 (2016)

\bibitem{ancona2021rml}
Ancona, D., Franceschini, L., Ferrando, A., Mascardi, V.: {RML}: theory and practice of a domain specific language for runtime verification. Science of Computer Programming  \textbf{205} (2021)

\bibitem{bester2023counting}
Bester, T., Rosman, B., James, S., Tasse, G.N.: Counting reward automata: Sample efficient reinforcement learning through the exploitation of reward function structure. arXiv preprint arXiv:2312.11364  (2023)

\bibitem{de2020restraining}
De~Giacomo, G., Iocchi, L., Favorito, M., Patrizi, F.: Restraining bolts for reinforcement learning agents. In: Proceedings of the AAAI Conference on Artificial Intelligence. vol.~34, pp. 13659--13662 (2020)

\bibitem{dietterich2000hierarchical}
Dietterich, T.G.: Hierarchical reinforcement learning with the {MAXQ} value function decomposition. Journal of artificial intelligence research  \textbf{13},  227--303 (2000)

\bibitem{falcone2013tutorial}
Falcone, Y., Havelund, K., Reger, G.: A tutorial on runtime verification. Engineering dependable software systems pp. 141--175 (2013)

\bibitem{fu2014probably}
Fu, J., Topcu, U.: Probably approximately correct {MDP} learning and control with temporal logic constraints. arXiv preprint arXiv:1404.7073  (2014)

\bibitem{icarte2018using}
Icarte, R.T., Klassen, T., Valenzano, R., McIlraith, S.: Using reward machines for high-level task specification and decomposition in reinforcement learning. In: International Conference on Machine Learning. pp. 2107--2116. PMLR (2018)

\bibitem{icarte2022reward}
Icarte, R.T., Klassen, T.Q., Valenzano, R., McIlraith, S.A.: Reward machines: Exploiting reward function structure in reinforcement learning. Journal of Artificial Intelligence Research  \textbf{73},  173--208 (2022)

\bibitem{jothimurugan2019composable}
Jothimurugan, K., Alur, R., Bastani, O.: A composable specification language for reinforcement learning tasks. Advances in Neural Information Processing Systems  \textbf{32} (2019)

\bibitem{leike2017ai}
Leike, J., Martic, M., Krakovna, V., Ortega, P.A., Everitt, T., Lefrancq, A., Orseau, L., Legg, S.: {AI} safety gridworlds. arXiv preprint arXiv:1711.09883  (2017)

\bibitem{levina2024numeric}
Levina, K., Pappas, N., Karapantelakis, A., Feljan, A.V., Seipp, J.: Numeric reward machines. arXiv preprint arXiv:2404.19370  (2024)

\bibitem{li2017reinforcement}
Li, X., Vasile, C.I., Belta, C.: Reinforcement learning with temporal logic rewards. In: 2017 IEEE/RSJ International Conference on Intelligent Robots and Systems (IROS). pp. 3834--3839. IEEE (2017)

\bibitem{ng1999policy}
Ng, A.Y., Harada, D., Russell, S.: Policy invariance under reward transformations: Theory and application to reward shaping. In: ICML. vol.~99, pp. 278--287 (1999)

\bibitem{parr1997reinforcement}
Parr, R., Russell, S.: Reinforcement learning with hierarchies of machines. Advances in neural information processing systems  \textbf{10} (1997)

\bibitem{sadigh2014learning}
Sadigh, D., Kim, E.S., Coogan, S., Sastry, S.S., Seshia, S.A.: A learning based approach to control synthesis of markov decision processes for linear temporal logic specifications. In: 53rd IEEE Conference on Decision and Control. pp. 1091--1096. IEEE (2014)

\bibitem{sutton2018reinforcement}
Sutton, R.S., Barto, A.G.: Reinforcement learning: An introduction. MIT press (2018)

\bibitem{sutton1999between}
Sutton, R.S., Precup, D., Singh, S.: Between {MDP}s and semi-{MDP}s: A framework for temporal abstraction in reinforcement learning. Artificial intelligence  \textbf{112}(1-2),  181--211 (1999)

\bibitem{unniyankal2023rmlgym}
Unniyankal, H., Belardinelli, F., Ferrando, A., Malvone, V.: {RMLG}ym: a formal reward machine framework for reinforcement learning. In: WOA. pp. 1--16 (2023)

\bibitem{varricchione2025pushdown}
Varricchione, G., Klassen, T.Q., Alechina, N., Dastani, M., Logan, B., McIlraith, S.A.: Pushdown reward machines for reinforcement learning. arXiv preprint arXiv:2508.06894  (2025)

\bibitem{watkins1992q}
Watkins, C.J., Dayan, P.: Q-learning. Machine learning  \textbf{8},  279--292 (1992)

\end{thebibliography}

%
%
%
%

\newpage

\appendix

\section{Experiment Details}

\subsection{Numerical Experiment - Flexibility Comparison}

During this experiment, for each value of $N$ a method was tested on 20 iterations were run, allowing us to gather the mean and standard deviation which are reported in the main text. The hyperparameters used in the experiment are shown in Figure~\ref{hyperparameters}. The same hyperparameters were used for all of the methods and were chosen based on performance during initial testing.

\begin{figure}[h]
\begin{itemize}
    \item Learning Rate: 0.5
    \item Initial Epsilon:  0.4  
    \item Epsilon Decay: 0.99
    \item Discount Factor: 0.9
\end{itemize}
\caption{Hyperparameters}
\label{hyperparameters}
\end{figure}

At each step of the task RML Reward Machines output a verdict. The majority of steps would be given a Currently False verdict. Successful completion of the task would lead to a Currently True verdict, while failure would lead to a False verdict. True verdicts would not be observed during this task.

Rewards are assigned based on the verdict: Currently True verdicts grant a reward of 100, Currently False verdicts grant 0, and False verdicts grant -40. On top of these rewards a +10 reward is given each time the monitor state changes. A small reward of +2 is also given the first time a monitor state-environment state pair is observed, encouraging exploration.

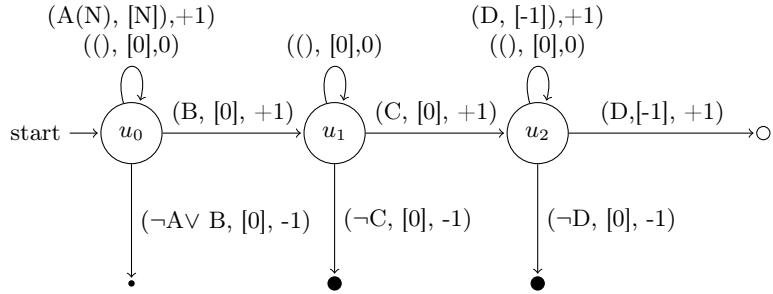
\begin{figure}[h]
    \centering
\begin{tikzpicture}[shorten >=1pt, node distance=2.7cm, on grid, auto]
   \node[state, initial] (q_0)   {$u_0$};
   \node[state] (q_1) [right=of q_0] {$u_1$};
   \node[state] (q_2) [right=of q_1] {$u_2$};
   
   \node[draw, circle, fill=black, minimum size=2pt, inner sep=0pt] (failure_0) [below=2cm of q_0] {};
   \node[draw, circle, fill=black, minimum size=5pt, inner sep=0pt] (failure_1) [below=2cm of q_1] {};
   \node[draw, circle, fill=black, minimum size=5pt, inner sep=0pt] (failure_2) [below=2cm of q_2] {};

   \node[draw, circle, fill=white, minimum size=5pt, inner sep=0pt] (success) [right=3cm of q_2] {};
   
   \path[->]
    (q_0) edge [above] node {(B, [0], +1)} (q_1)
    (q_1) edge [above] node {(C, [0], +1)} (q_2)
    (q_2) edge [above] node {(D,[-1], +1)} (success)
    (q_0) edge [loop above] node[align=center] {(A(N), [N]),+1) \\ ((), [0],0)} (q_0)
    (q_1) edge [loop above] node {((), [0],0)} (q_1) 
    (q_2) edge [loop above] node[align=center] {(D, [-1]),+1) \\ ((), [0],0)} (q_2)

    (q_0) edge [below, right] node {($\neg$A$\vee$ B, [0], -1)} (failure_0)
    (q_1) edge [below, right] node {($\neg$C, [0], -1)} (failure_1)
    (q_2) edge [below, right] node {($\neg$D, [0], -1)} (failure_2);
\end{tikzpicture}
\caption{Counting Reward Automata for Numerical Experiment}
\label{fig:numerical_experiment_cra}
\end{figure}

The Counting Reward Automata designed for the experiment is shown in Figure~\ref{fig:numerical_experiment_cra}. The format of the transitions is (observation,[increment], reward). For example, (A(1),[1],+1) would be the relevant transition when the value 1 is observed for A. The () observation represents when the agent is on a blank square. The value of the counter is omitted from the graph. On the final node, the loop transition is used if the stored count is greater than 0 when D is observed, otherwise the transition to the white success node is used. Each transition that advances through the task is provided a +1 reward. This includes transitions between states, as well as when A and D are observed but the state remains the same. Transitions that lead to task failure are given -1 reward. Similar to RML Reward Machines, CQL receive a small exploration reward of +0.1 when machine state-environment state pairs are observed for the first time.

\subsection{Numerical Experiment - Effect of Monitor State Visibility Details}

The RML Reward Machine-based approach used the same hyperparameters as in the previous experiment, demonstrating its flexibility relative to CRA. These hyperparameters are listed in Figure~\ref{hyperparameters} in Appendix B.1.

The hyperparameters used by RMLGym are shown in Figure~\ref{hyperparameters_rmlgym}. They were selected via a grid search on the same task, with the best-performing configuration used.

\begin{figure}[h]
\begin{itemize}
    \item Learning Rate: 0.01
    \item Initial Epsilon:  0.75  
    \item Epsilon Decay: 0.999
    \item Discount Factor: 0.9
\end{itemize}
\caption{RMLGym Hyperparameters}
\label{hyperparameters_rmlgym}
\end{figure}

\end{document}


\appendix
\section*{Appendix}

\section{RML Term Syntax}

\begin{figure}[ht]
\centering
\begin{align*}
\text{RS} & ::= \text{Main} = \text{RE}; \text{RD}_1; \ldots; \text{RD}_n; \text{    (RML Specification)}\\
\text{RD} & ::= \text{Id} = \text{RE} \mid \text{Id}(\overline{x}) = \text{RE} \quad \quad \text{     
   (RML Definition)} \\
\text{rb} & ::= x \mid l \quad (\text{RML basic data expression}) \\
\theta & ::= \lambda (\text{rb}_1, \ldots, \text{rb}_n) \quad (\text{RML event type pattern}) \\
\text{rd} & ::= \text{rb} \mid u \text{ rd} \mid \text{rd}_1 \, o \, \text{rd}_2 \quad (\text{RML data expression}) \\
\text{RE} & ::= \text{empty} \, (\text{empty trace}) \\
& \mid \text{all} \, (\text{universe}) \\
& \mid \text{none} \, (\text{none}) \\
& \mid \theta \, (\text{event type pattern}) \\
& \mid \text{Id} \, (\text{specification name}) \\
& \mid \text{Id}(\text{rd}_1, \ldots, \text{rd}_n) \, (\text{generic instantiation}) \\
& \mid \text{RE}_1 \, \text{RE}_2 \, (\text{concatenation}) \\
& \mid \text{RE}_1 \land \text{RE}_2 \, (\text{intersection}) \\
& \mid \text{RE}_1 \lor \text{RE}_2 \, (\text{union}) \\
& \mid \text{RE}_1 \mid \text{RE}_2 \, (\text{shuffle}) \\
& \mid \{\text{let } x; \, \text{RE}\} \, (\text{parametric block}) \\
& \mid \text{RE}? \, (\text{optional}) \\
& \mid \text{RE}^+ \, (\text{postfix plus}) \\
& \mid \text{RE}^* \, (\text{Kleene star}) \\
& \mid \text{RE}! \, (\text{prefix closure}) \\
& \mid \theta \gg \text{RE}_1 : \text{RE}_2 \, (\text{conditional filter}) \\
& \mid \theta \gg \text{RE} \, (\text{unconditional filter}) \\
& \mid \text{if} \, (\text{rd}) \, \text{RE}_1 \, \text{else} \, \text{RE}_2 \, (\text{conditional expression})
\end{align*}
\caption{Full RML Syntax; RE terms are finite}
\label{RML_syntax}
\end{figure}

\newpage
\section{Experiment Details}

\subsection{Numerical Experiment - Flexibility Comparison}

During this experiment, for each value of $N$ a method was tested on 20 iterations were run, allowing us to gather the mean and standard deviation which are reported in the main text. The hyperparameters used in the experiment are shown in Figure~\ref{hyperparameters}. The same hyperparameters were used for all of the methods and were chosen based on performance during initial testing.

\begin{figure}[h]
\begin{itemize}
    \item Learning Rate: 0.5
    \item Initial Epsilon:  0.4  
    \item Epsilon Decay: 0.99
    \item Discount Factor: 0.9
\end{itemize}
\caption{Hyperparameters}
\label{hyperparameters}
\end{figure}

At each step of the task RML Reward Machines output a verdict. The majority of steps would be given a Currently False verdict. Successful completion of the task would lead to a Currently True verdict, while failure would lead to a False verdict. True verdicts would not be observed during this task.

Rewards are assigned based on the verdict: Currently True verdicts grant a reward of 100, Currently False verdicts grant 0, and False verdicts grant -40. On top of these rewards a +10 reward is given each time the monitor state changes. A small reward of +2 is also given the first time a monitor state-environment state pair is observed, encouraging exploration.

The Counting Reward Automata designed for the experiment is shown in Figure~\ref{fig:numerical_experiment_cra}. The format of the transitions is (observation,[increment], reward). For example, (A(1),[1],+1) would be the relevant transition when the value 1 is observed for A. The () observation represents when the agent is on a blank square. The value of the counter is omitted from the graph. On the final node, the loop transition is used if the stored count is greater than 0 when D is observed, otherwise the transition to the white success node is used. Each transition that advances through the task is provided a +1 reward. This includes transitions between states, as well as when A and D are observed but the state remains the same. Transitions that lead to task failure are given -1 reward. Similar to RML Reward Machines, CQL receive a small exploration reward of +0.1 when machine state-environment state pairs are observed for the first time.

\begin{figure}[h]
    \centering
\begin{tikzpicture}[shorten >=1pt, node distance=2.7cm, on grid, auto]
   \node[state, initial] (q_0)   {$u_0$};
   \node[state] (q_1) [right=of q_0] {$u_1$};
   \node[state] (q_2) [right=of q_1] {$u_2$};
   
   \node[draw, circle, fill=black, minimum size=2pt, inner sep=0pt] (failure_0) [below=2cm of q_0] {};
   \node[draw, circle, fill=black, minimum size=5pt, inner sep=0pt] (failure_1) [below=2cm of q_1] {};
   \node[draw, circle, fill=black, minimum size=5pt, inner sep=0pt] (failure_2) [below=2cm of q_2] {};

   \node[draw, circle, fill=white, minimum size=5pt, inner sep=0pt] (success) [right=3cm of q_2] {};
   
   \path[->]
    (q_0) edge [above] node {(B, [0], +1)} (q_1)
    (q_1) edge [above] node {(C, [0], +1)} (q_2)
    (q_2) edge [above] node {(D,[-1], +1)} (success)
    (q_0) edge [loop above] node[align=center] {(A(N), [N]),+1) \\ ((), [0],0)} (q_0)
    (q_1) edge [loop above] node {((), [0],0)} (q_1) 
    (q_2) edge [loop above] node[align=center] {(D, [-1]),+1) \\ ((), [0],0)} (q_2)

    (q_0) edge [below, right] node {($\neg$A$\vee$ B, [0], -1)} (failure_0)
    (q_1) edge [below, right] node {($\neg$C, [0], -1)} (failure_1)
    (q_2) edge [below, right] node {($\neg$D, [0], -1)} (failure_2);
\end{tikzpicture}
\caption{Counting Reward Automata for Numerical Experiment}
\label{fig:numerical_experiment_cra}
\end{figure}

\subsection{Numerical Experiment - Effect of Monitor State Visibility Details}

The RML Reward Machine-based approach used the same hyperparameters as in the previous experiment, demonstrating its flexibility relative to CRA. These hyperparameters are listed in Figure~\ref{hyperparameters} in Appendix B.1.

The hyperparameters used by RMLGym are shown in Figure~\ref{hyperparameters_rmlgym}. They were selected via a grid search on the same task, with the best-performing configuration used.

\begin{figure}[h]
\begin{itemize}
    \item Learning Rate: 0.01
    \item Initial Epsilon:  0.75  
    \item Epsilon Decay: 0.999
    \item Discount Factor: 0.9
\end{itemize}
\caption{RMLGym Hyperparameters}
\label{hyperparameters_rmlgym}
\end{figure}